\documentclass{article}
\usepackage{iclr2027_conference,times}

\usepackage{amsmath,amsfonts,bm}

\def\eqref#1{equation~\ref{#1}}

\def\1{\bm{1}}

\DeclareMathAlphabet{\mathsfit}{\encodingdefault}{\sfdefault}{m}{sl}
\SetMathAlphabet{\mathsfit}{bold}{\encodingdefault}{\sfdefault}{bx}{n}

\usepackage{hyperref}
\usepackage{url}
\usepackage{graphicx}
\usepackage{booktabs}
\usepackage{xcolor}
\usepackage{caption}
\usepackage{makecell}

\newcommand{\bench}{\textsc{BigDetail}}
\newcommand{\rev}[1]{#1}

\title{NOFACE3D: NOise-scheduled Focused Anchoring and Controlled Enhancement for 3D Generation}

\author{James Rose$^{1*}$ \quad Jingwen Liang$^{1*}$ \quad Wenchao Ma$^{1,2}$ \quad Oren Jacob$^{1}$ \quad Haomiao Jiang$^{1}$ \\
{\normalfont $^{1}$Roblox \qquad $^{2}$The Pennsylvania State University}
}

\iclrfinalcopy
\begin{document}

\maketitle
\lhead{Preprint}
\begingroup\renewcommand\thefootnote{\fnsymbol{footnote}}\footnotetext[1]{Equal contribution.}\endgroup

\begin{center}
  \includegraphics[width=\linewidth]{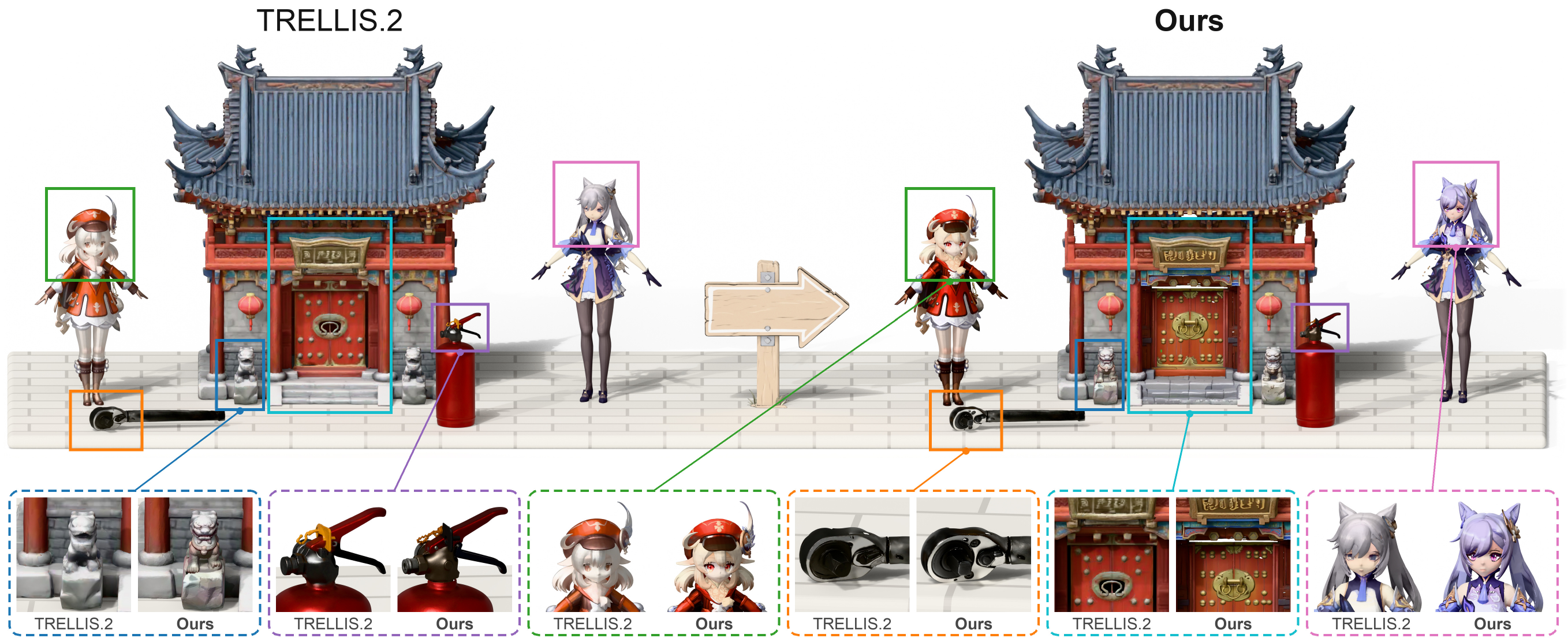}
  \captionof{figure}{\textbf{\textsc{NOFACE3D}: training-free localized refinement
  of image-to-3D assets.} Left: a scene of assets, each generated from a single
  image by TRELLIS.2. Right: the same assets after our method refines the boxed
  regions, preserving the rest of each asset. Dashed panels show enlarged
  comparisons: the base generation leaves fine details soft, while ours
  recovers them with sharp geometry and texture, with no extra training or
  supervision.}
  \label{fig:teaser}
\end{center}

\begin{abstract}
Recent image-to-3D models generate complete, high-quality assets from a single
image, with global shape and appearance that are already convincing; what
remains is the fine scale, where small, thin, elongated, or detail-rich regions
are reconstructed softly and under-resolved. This gap can arise in generators
with fixed-resolution spatial grids: the same spatial sampling density is used
for a large, smooth, low-curvature surface and for a thin, high-curvature
structure that carries most of the perceived detail. To this end, we introduce
NOFACE3D, a training-free framework that refines these detail-critical regions
while preserving the rest of the asset. Given a generated mesh, its input
image, and a refinement box, NOFACE3D extracts the selected submesh, scales it
up to fill the generator's input volume, and re-encodes it into the
full-resolution latent grid. Conditioned on the image crop, it refines the
region through flow-based sampling with a spatially varying noise schedule
that anchors boundary latents. The refined region is then integrated back
into the original asset, preserving the surrounding geometry and appearance.
We implement NOFACE3D on TRELLIS.2 and its sparse
voxel representation; all of it operates purely at inference time on the
frozen pretrained generator. We evaluate NOFACE3D on PAniC-3D, TexVerse and \bench{}, a new benchmark of $339$ assets with annotated detail-critical regions. On all three, it generates sharper geometry and finer texture detail than the baselines. Project page: \href{https://idkcouscous.github.io/NoFACE3D/}{\textcolor{blue}{https://idkcouscous.github.io/NoFACE3D/}}
\end{abstract}

\section{Introduction}

Single-image 3D generation now produces complete, textured assets in seconds.
Feed-forward reconstruction models predict 3D representations directly from
image features \citep{hong2024lrm,tang2025lgm,xu2024instantmesh}, while native 3D
latent generative models trained on large asset collections
\citep{deitke2023objaverse} synthesize objects with convincing proportions,
silhouettes, and appearance
\citep{zhang2024clay,li2025triposg,zhao2025hunyuan3d2}. Structured-latent
generators such as TRELLIS and TRELLIS.2
\citep{xiang2024trellis,xiang2025trellis2} attach a local code to each occupied
voxel of a sparse grid. Their output is already convincing on large,
low-curvature surfaces.

However, fine details such as fingers, springs, and engravings often
appear soft or blurred.
This gap can arise from how spatial capacity is allocated: a fixed-resolution
grid provides the same sampling density to large, smooth surfaces and small,
high-curvature structures.
For example, one full-body character generated by TRELLIS.2
\citep{xiang2025trellis2} occupies approximately $4{,}000$ voxels in the $64^3$
shape-latent grid ($1.5\%$ of its cells), with only about $1{,}400$ voxels
($0.5\%$) representing the head crop.
The challenge is therefore to give detail-critical regions more spatial
capacity while preserving the parts that are already well reconstructed.

Increasing the global grid resolution offers finer spatial sampling, but also
increases computational cost across the entire asset, even when only a small
region needs refinement \citep{wu2025direct3ds2}. Refiners trained on synthetic
mesh degradation may not fully address detail lost through limited spatial
sampling during generation \citep{deng2024detailgen3d}. Render-space enhancement
can improve appearance, but does not directly increase the latent capacity
allocated to a selected 3D region \citep{ryu2025elevate3d}.

To address this challenge, we propose \textsc{NOFACE3D}, a training-free
framework built on a frozen image-to-3D generator. We extract the target
submesh, rescale it to the
generator's input cube, and re-encode it with the pretrained encoders.
The generator's full-resolution latent grid now represents this region alone,
increasing its effective resolution in the original asset's coordinates.
The input image crop guides regeneration.

A spatially varying noise schedule anchors latents near attachment boundaries,
with stronger regeneration farther away. Finally, we map the
refined region back to the original coordinate frame and combine it with the
surrounding mesh.

Our evaluation focuses on whether refinement produces sharper geometry and
finer texture detail while preserving overall asset quality. We evaluate
\textsc{NOFACE3D} on PAniC-3D and TexVerse, and introduce \bench{} to assess
localized refinement on diverse objects with detail-critical regions annotated
in 2D and 3D. Where artist ground truth is available, we
report appearance metrics and pointwise geometry measures. We interpret
Chamfer distance and F-score as geometry checks, since they are dominated by
large, smooth surfaces \citep{wang2018pixel,chen2025dora} and may not reflect
fine detail that artist references carry in textures rather than geometry.
To assess visible improvements in geometry and texture, we use qualitative
comparisons and GPTEval3D evaluations across all three datasets, alongside a
human preference study on PAniC-3D. Across these evaluations,
\textsc{NOFACE3D} outperforms the original TRELLIS.2 results and the refinement
and editing baselines.

Our contributions are as follows.
\begin{itemize}
  \item \textsc{NOFACE3D}, a training-free framework for localized refinement
  that scales up a selected region to fill the generator's input volume and
  re-encodes it into the full-resolution latent grid.
  \item A spatially varying noise schedule for flow-based sampling that anchors
  boundary latents and increases refinement strength with distance from
  attachment boundaries.
  \item \bench{}, a benchmark of $339$ assets reconstructed from generated
  images, with detail-critical regions annotated in both 2D and 3D.
  Qualitative comparisons and GPTEval3D evaluations on PAniC-3D, TexVerse,
  and \bench{} demonstrate sharper geometry and finer texture detail.
\end{itemize}

\section{Related Work}
\subsection{Text- and image-to-3D generation}
Early text-to-3D methods lift 2D diffusion priors into 3D by score distillation,
optimizing a per-scene representation so its renderings match a pretrained image
model \citep{poole2023dreamfusion,lin2023magic3d,wang2023prolificdreamer,chen2023fantasia3d};
they are flexible but slow, prone to view inconsistency, and tend toward
oversmoothed geometry, which multi-view diffusion mitigates by generating several
consistent views before lifting
\citep{liu2023zero123,shi2024mvdream,long2024wonder3d}. Feed-forward large
reconstruction models instead predict a 3D representation in a single pass,
trading per-scene optimization for speed
\citep{hong2024lrm,li2024instant3d,tang2025lgm,xu2024instantmesh}. The now
dominant family trains native 3D latent generators on large asset collections,
using compact vector-set latents
\citep{zhang2023shape2vecset,zhao2024michelangelo,zhang2024clay,li2025triposg,zhao2025hunyuan3d2}
or structured sparse-voxel latents \citep{xiang2024trellis,xiang2025trellis2}. Because
dense volumetric capacity scales cubically, recent work attacks this cost with
sparse-volume attention \citep{wu2025direct3ds2} and non-uniform representations
that place resolution where occupancy is
\citep{xiong2025octfusion,ren2024xcube}, while part-based generators disperse
capacity across semantic parts so each is generated at higher effective
resolution
\citep{lin2025partcrafter,yang2025holopart,chen2025omnipart,sun2026eipart}. A
parallel autoregressive line generates meshes as triangle token sequences,
targeting artist-like topology directly
\citep{siddiqui2024meshgpt,chen2024meshanything,tang2024edgerunner}.
\textsc{NOFACE3D} shares the motivation of the capacity-aware methods, but instead
of redesigning the representation or retraining, it reallocates the existing
budget at inference time to a single semantic region.

\subsection{Mesh and geometry refinement}
A dedicated line enhances the geometry of an existing coarse asset. DetailGen3D
models coarse-to-fine enhancement as a data-dependent latent flow with token
matching, trained on synthetically degraded pairs \citep{deng2024detailgen3d}.
Elevate3D refines view by view, alternating a high-frequency-preserving SDEdit
\citep{meng2022sdedit} variant for texture with monocular-geometry updates for shape, explicitly
targeting the trade-off a single global noise level imposes
\citep{ryu2025elevate3d}. MeshReGen unifies enhancement, reconstruction, and
editing as one regeneration problem and reports that re-encoding degraded
geometry into a structured latent and resampling does not recover fine detail
\citep{park2026meshregen}. A related line bridges through predicted normals to raise
detail fidelity before geometry synthesis \citep{ye2025hi3dgen}, while
no-ground-truth surface metrics \citep{lee2026revive3d} and region-annotated
character benchmarks \citep{chen2023panic3d} support scoring such refinements.
\textsc{NOFACE3D} shares this goal but is training-free and
localized: it learns no degradation-specific mapping, operates on a frozen
generator, and refines one region at increased effective resolution while
leaving the geometry outside it essentially unchanged.

\section{Method}
\label{sec:method}

\textsc{NOFACE3D} takes a generated asset, its conditioning image, and a bounding box
marking a detail-critical region, and regenerates only that region with the
frozen generator. The region is scaled up to fill the generator's input volume
and re-encoded into its full-resolution latent grid
(Section~\ref{sec:reencode}); a boundary-distance field controls
the noise strength and release time of each latent token
(Section~\ref{sec:field}); geometry and texture are resampled under that field
(Sections~\ref{sec:sampling},~\ref{sec:texture}); and the result is cut back
into the asset in mesh space (Section~\ref{sec:rejoin}). Implementation details
and hyperparameters for every step are in Appendix~\ref{app:impl}.

\begin{figure}
    \centering
    \includegraphics[width=\linewidth]{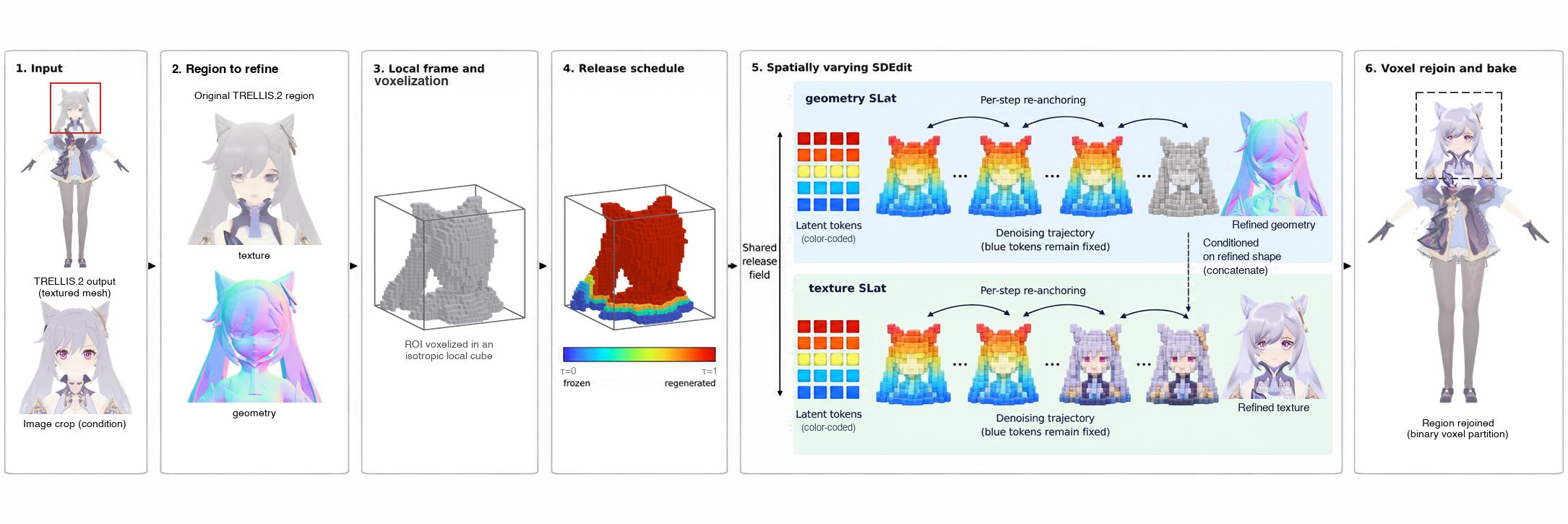}
    \vspace{-6mm}
    \caption{\textbf{Overview of \textsc{NOFACE3D}.}  Given an asset generated by an
off-the-shelf image-to-3D model and a user-marked region, we scale up the region
to fill the generator's input volume and re-encode it into the full-resolution
latent grid, assign each token
a release level $\tau$ from a boundary-distance field (anchored near attachment
boundaries, with stronger regeneration farther away), and resample
geometry and texture with spatially varying SDEdit \citep{meng2022sdedit}, re-anchoring unreleased
tokens to the base asset at every step. The refined mesh replaces the base
faces inside the crop box.}
\label{fig:pipeline}
\end{figure}

\subsection{Preliminaries}
\label{sec:prelim}

We build on TRELLIS.2 \citep{xiang2025trellis2}, which first predicts a sparse
structure (SS) of active voxels, then generates separate structured latents
(SLats) for shape and texture. The asset is represented in a fixed cube
$\mathcal{C} = [-\tfrac12, \tfrac12]^3$, and generation uses flow models
conditioned on DINOv3 \citep{simeoni2025dinov3} features of one image. Our inference pipeline comprises
four sampling stages: sparse structure (SS), low-resolution shape,
high-resolution shape, and texture. The SS model samples a dense
$16^3$ latent that decodes to coarse occupancy; shape
SLats are then sampled on a sparse $32^3$ support and a sparse $64^3$ grid, and
a texture SLat on the same $64^3$ coordinates. The decoders lift the shape and
texture SLats to geometry and PBR attributes on $1024^3$ voxels, from which a
textured mesh is extracted. Each stage follows the
flow-matching path \citep{lipman2023flow,liu2023rectified}
\begin{equation}
  x_t = (1 - t)\, x_0 + \big(\sigma_{\min} + (1 - \sigma_{\min})\, t\big)\, \epsilon,
  \qquad \epsilon \sim \mathcal{N}(0, I),
  \label{eq:path}
\end{equation}
where $x_0$ is a clean latent, $t \in [0,1]$ controls the noise level, and
$\sigma_{\min}$ is the residual noise scale at $t=0$. At $t=1$, $x_t$ is pure
noise. Ordinary generation integrates the learned velocity $v_\theta$ from
$t=1$ to $0$ using Euler steps. We reuse the
pretrained models in a cascaded inference pipeline to refine the selected
region after rescaling and re-encoding it.

\subsection{Region selection and local re-encoding}
\label{sec:reencode}

The refinement region is specified by a 3D bounding box $B$, with a
corresponding 2D bounding box $\Omega$ in the conditioning image. The 3D box
can be supplied directly, as in \bench{}, or
derived from the image region. We extract the submesh selected by $B$ without
splitting triangles, then center it and scale it uniformly to fit the
generator's input cube, preserving its proportions.

Only this rescaled submesh is converted to the generator's $1024^3$ sparse
voxel representation and re-encoded with the pretrained encoders to obtain the
base latents for each stage. Rescaling increases the region's effective spatial
resolution across all latent stages and the decoded voxel representation.
Regeneration is conditioned on the crop of the input image
within $\Omega$, passed through the generator's standard preprocessing.

\subsection{Boundary-distance release field}
\label{sec:field}

Our refinement requires at least one \emph{attachment boundary}: a crop-box
face crossed by the base mesh where the selected region connects to the
surrounding asset. We anchor nearby
latents to the base asset and allow stronger regeneration farther away.
Each latent token receives a \emph{release level}
$\tau \in [0, \tau_{\max}]$. The same $\tau$ sets its initial noising time
$t=\tau$ and determines when it stops following the noised base latent and
follows the learned flow instead (Section~\ref{sec:sampling}). Larger $\tau$
means stronger initial noising and earlier release. Tokens with $\tau=0$
remain anchored and are restored to their base values at the end; tokens with
$\tau=1$ start from pure noise. The per-stage maximum $\tau_{\max}\le1$
controls refinement strength.

For a token at position $u$, we measure its distance to each attachment boundary
as a fraction of the box extent in that direction, and take the smallest value
$d(u)$. We set $\tau=0$ within an \emph{anchored band}
of width $g$, then increase it smoothly over a \emph{transition band} of width
$r$:
\begin{equation}
  \tau(u) = \tau_{\max}\; \rho^2 (3 - 2 \rho),
  \qquad
  \rho = \mathrm{clamp}\!\Big( \frac{d(u) - g}{r},\, 0,\, 1 \Big).
  \label{eq:field}
\end{equation}
Beyond the transition band ($d(u)\ge g+r$), $\tau$ reaches $\tau_{\max}$.

Each stage uses this distance-based rule with its own empirically chosen band
widths $(g,r)$. We choose the widths for sparse
structure (SS), shape, and texture generation to satisfy
\begin{equation}
  g_{\mathrm{shape}} + r_{\mathrm{shape}} \le g_{\mathrm{SS}},
  \qquad
  g_{\mathrm{tex}} + r_{\mathrm{tex}} \le g_{\mathrm{shape}},
  \label{eq:nesting}
\end{equation}
so each group's anchored and transition bands lie within the preceding group's
anchored band.

\subsection{Spatially varying flow sampling}
\label{sec:sampling}

Let $z$ be the base latent for a given stage and $\bar{z}$ be its per-channel
standardized version (the SS latent is used as is), following the TRELLIS.2
implementation \citep{xiang2025trellis2}. The release level $\tau_i$ for token
$i$ is given by Eq.~\ref{eq:field}. We sample Gaussian noise $\epsilon$ once
per stage and reuse it throughout that stage's sampling. Using
Eq.~\ref{eq:path}, we obtain the noised base latent of token $i$ at time $t$ as
\[
  x^{\mathrm{src}}_i(t) = (1 - t)\, \bar{z}_i
  + \big(\sigma_{\min} + (1 - \sigma_{\min})\, t\big)\, \epsilon_i.
\]

Sampling runs over a descending time ladder $t_0 > t_1 > \dots > t_K=0$;
its construction differs between SS and SLat stages.
Each token is initialized at its own level,
$x_i(t_0)=x^{\mathrm{src}}_i(\tau_i)$, which need not match the first model
time $t_0$. After
each Euler step
$\tilde{x} = x(t_k) - (t_k - t_{k+1})\, v_\theta(x(t_k), t_k)$, with the image
condition and classifier-free guidance \citep{ho2022cfg} of the generator, unreleased tokens
are reset to the noised base latent at the next sampling time.
We call this operation \emph{re-anchoring}:
\begin{equation}
  x_i(t_{k+1}) =
  \begin{cases}
    x^{\mathrm{src}}_i(t_{k+1}) & \text{if } \tau_i < t_{k+1}, \\
    \tilde{x}_i  & \text{otherwise},
  \end{cases}
  \quad k < K - 1,
  \qquad
  x_i(t_K) =
  \begin{cases}
    \bar{z}_i   & \text{if } \tau_i < t_{K-1}, \\
    \tilde{x}_i & \text{otherwise}.
  \end{cases}
  \label{eq:anchor}
\end{equation}
Through self-attention, the flow model can use unreleased tokens as base-asset
context at the corresponding noise level when updating released tokens.
At the end of sampling, tokens that were never released, including those in
the anchored band, are restored exactly to their base latents.
The construction is the structured-latent counterpart of
per-pixel edit strength in 2D diffusion
\citep{levin2023differential,kim2024rad}; it shares the known-region
replacement of RePaint \citep{lugmayr2022repaint} but uses a graded release
instead of a binary mask and performs no resampling jumps.

The re-anchored sampling of Eq.~\ref{eq:anchor} runs at every stage of the shape
cascade with $\tau$ evaluated at that stage's coordinates, so coarse structure
changes only where it is released. We retain the generator's guidance and
$\sigma_{\min}$ for these stages. The shape SLats
additionally blend their output toward the base latent by release level, and
the SS stage copies the base occupancy wherever $\tau$ vanishes.

\subsection{Texture refinement}
\label{sec:texture}

Texture is refined after geometry, on the refined shape coordinates; the
decoded refined mesh is never re-encoded. The base texture SLat is encoded in
the same local frame from the crop seed and the base asset's PBR attributes,
and the texture model uses the same boundary-distance field, evaluated with the
texture-stage $g$ and $r$, and the same re-anchored sampling of
Eq.~\ref{eq:anchor}. By Eq.~\ref{eq:nesting}, the texture anchored and transition
bands lie inside the shape SLat's anchored band, where base texture values are
available at matching coordinates. The result decodes to PBR channels on the refined
$1024^3$ voxels.

\subsection{Reintegration in mesh space}
\label{sec:rejoin}

The refined region is decoded to a mesh with per-vertex PBR values and mapped
back to the original coordinate frame using the inverse transforms. This restores the region's position, orientation and
scale without an additional alignment fit. The cut is
made in mesh space with the same crop box $B$ used for selection: base faces
whose centroid $c_f$ lies inside $B$ are removed, and the refined mesh takes
their place,
\begin{equation}
  \mathcal{M} =
  \{ f \in \mathcal{M}_{\mathrm{base}} : c_f \notin B \}
  \;\cup\;
  \mathcal{M}_{\mathrm{ref}}.
  \label{eq:rejoin}
\end{equation}
The two parts are concatenated, with no mesh boolean, vertex welding or
stitching. The cut is binary although $\tau$ is smooth: the transition band
controls how strongly latents are released, while anchoring encourages compatible geometry
and a smooth transition in color at attachment boundaries. The joined mesh is then re-voxelized at 1024³ and remeshed for export. 

\begin{table}[t]
\centering
\caption{Comparison with refinement and editing methods. Whole-asset metrics against GT; CD and SNE ${\times}100$; IoU at $128^3$; FID is set-level. GPT-5.6: pairwise win share (1--10, chance $5.0$) for overall quality and geometry detail.}
\label{tab:baselines}
\footnotesize
\setlength{\tabcolsep}{2pt}
\begin{tabular*}{\textwidth}{@{\extracolsep{\fill}} l ccccc ccccc cc}
\toprule
& \multicolumn{5}{c}{Geometry} & \multicolumn{5}{c}{Appearance}
& \multicolumn{2}{c}{GPT-5.6} \\
\cmidrule(lr){2-6} \cmidrule(lr){7-11} \cmidrule(lr){12-13}
& CD & F1$_{0.25\%}$ & F1$_{1\%}$ & IoU & SNE & PSNR & SSIM & LPIPS & DINO & FID
& Ovr. & Geo. \\
Method & $\downarrow$ & $\uparrow$ & $\uparrow$ & $\uparrow$ & $\downarrow$
       & $\uparrow$ & $\uparrow$ & $\downarrow$ & $\uparrow$ & $\downarrow$
       & $\uparrow$ & $\uparrow$ \\
\midrule
\multicolumn{13}{l}{\emph{PAniC-3D} ($N{=}329$)} \\
GT              & \multicolumn{10}{c}{\emph{reference}} & $8.52$ & $6.83$ \\
TRELLIS.2       & $\mathbf{2.938}$ & $0.138$ & $0.389$ & $0.383$ & $1.12$ & $15.80$ & $0.882$ & $0.118$ & $0.792$ & $43.17$ & $6.48$ & $6.82$ \\
Easy3E          & $3.955$ & $0.074$ & $0.265$ & $0.294$ & $1.24$ & $13.17$ & $0.878$ & $0.128$ & $0.588$ & $98.74$ & $2.44$ & $3.02$ \\
Elevate3D       & $3.059$ & $0.115$ & $0.372$ & $0.377$ & $1.26$ & $16.59$ & $0.876$ & $0.122$ & $0.671$ & $68.25$ & $3.81$ & $0.71$ \\
DetailGen3D     & $2.939$ & $0.127$ & $\mathbf{0.393}$ & $0.386$ & $\mathbf{1.04}$ & -- & -- & -- & -- & -- & -- & $4.55$ \\
\textsc{NOFACE3D} & $2.939$ & $\mathbf{0.142}$ & $0.389$ & $\mathbf{0.391}$ & $1.12$ & $\mathbf{16.78}$ & $\mathbf{0.883}$ & $\mathbf{0.115}$ & $\mathbf{0.812}$ & $\mathbf{33.02}$ & $\mathbf{8.36}$ & $\mathbf{8.07}$ \\
\midrule
\multicolumn{13}{l}{\emph{TexVerse} ($N{=}151$)} \\
GT              & \multicolumn{10}{c}{\emph{reference}} & $7.30$ & $6.33$ \\
TRELLIS.2       & $7.27$ & $0.123$ & $0.306$ & $0.334$ & $1.97$ & $18.02$ & $0.860$ & $0.135$ & $0.728$ & $66.56$ & $3.84$ & $3.09$ \\
Easy3E          & $11.7$ & $0.060$ & $0.219$ & $0.250$ & $1.81$ & $17.25$ & $0.851$ & $0.145$ & $0.663$ & $74.87$ & $6.13$ & $5.61$ \\
Elevate3D       & $13.7$ & $0.058$ & $0.194$ & $0.198$ & $2.16$ & $16.97$ & $0.836$ & $0.158$ & $0.620$ & $73.94$ & $3.89$ & $3.14$ \\
DetailGen3D     & $35.4$ & $0.009$ & $0.032$ & $0.055$ & $\mathbf{1.23}$ & -- & -- & -- & -- & -- & -- & $5.32$ \\
\textsc{NOFACE3D} & $\mathbf{7.22}$ & $\mathbf{0.140}$ & $\mathbf{0.320}$ & $\mathbf{0.337}$ & $1.61$ & $\mathbf{19.57}$ & $\mathbf{0.863}$ & $\mathbf{0.134}$ & $\mathbf{0.741}$ & $\mathbf{64.92}$ & $\mathbf{7.22}$ & $\mathbf{6.50}$ \\
\bottomrule
\end{tabular*}
\vspace{-4mm}
\end{table}

\section{Experiments}
\label{sec:experiments}
We compare \textsc{NOFACE3D} with refinement and editing methods on PAniC-3D
and TexVerse, and with TRELLIS.2 on \bench{}, and in a human study
(Section~\ref{sec:baseline-results}). We then ablate the noise schedule
(Section~\ref{sec:ablation}) and show that the method also supports localized
3D editing (Section~\ref{sec:applications}).
\subsection{Setup}
\label{sec:setup}
\textbf{Datasets and baselines.}
\label{sec:bigdetail}
PAniC-3D \citep{chen2023panic3d} pairs single images of full-body characters
with artist meshes. Their head and upper body carry much of the perceived
identity but occupy only a small fraction of the latent grid, so the marked
region is a semantic crop of that part. We compare with
baselines on $329$ of the $4{,}892$ assets, because Elevate3D takes about
25 minutes per asset on an NVIDIA RTX A6000 \citep{ryu2025elevate3d}.
\textsc{NOFACE3D} runs the TRELLIS.2 cascade once per region, close to the cost
of one TRELLIS.2 generation, which takes about $17$\,s at $1024^3$ on an NVIDIA
H100 \citep{xiang2025trellis2}.
TexVerse \citep{zhang2025texverse} adds $151$ general
objects with authored PBR materials, none from Objaverse. We also propose
\bench{}, a benchmark of objects whose detail sits in a small region that a
uniform latent grid under-resolves. Each subject pairs a
large, smooth body with a compact cluster of named detail parts (e.g.\ a steel
tank with its valve and trigger assembly), spanning everyday and industrial
objects. We generate one image per subject with Nano Banana 2, reconstruct it
with the original TRELLIS.2 model, and annotators mark each detail-critical
region as an image rectangle and an oriented 3D box. The resulting $339$
assets have no artist mesh, so they are judged pairwise against the original
TRELLIS.2 result. The baselines, DetailGen3D
\citep{deng2024detailgen3d}, Elevate3D \citep{ryu2025elevate3d} and Easy3E
\citep{hu2026easy3e}, are each applied to the original TRELLIS.2 result.
\rev{We do not compare with VoxHammer \citep{li2025voxhammer}, a training-free
3D editing method, because it edits at the generator's native resolution and
does not increase it.}

\begin{figure}[t]
\centering
\includegraphics[width=\linewidth]{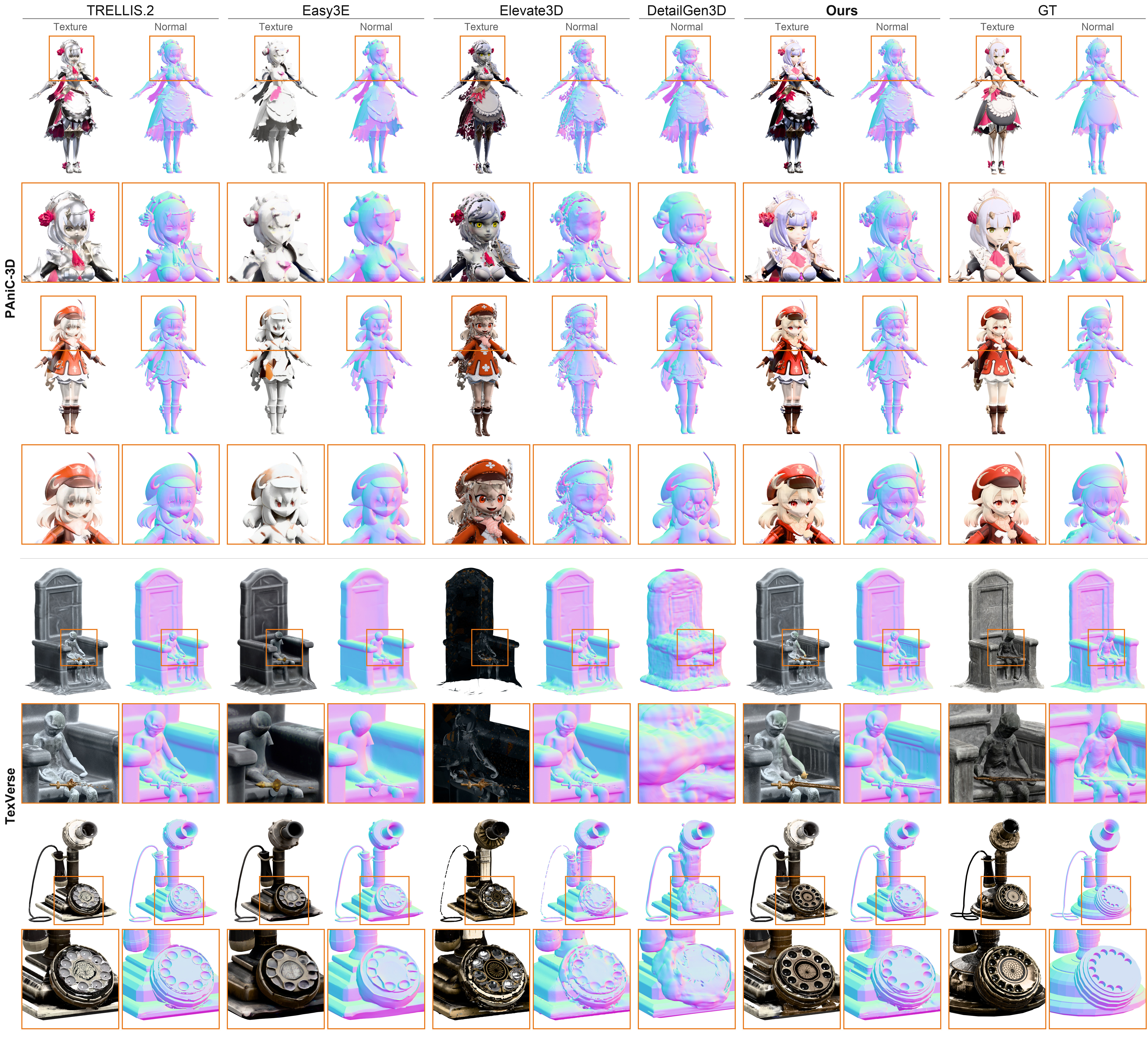}
\caption{\textbf{Qualitative comparison} on PAniC-3D (top) and TexVerse
(bottom): texture and normal map per method (DetailGen3D is untextured).}
\label{fig:qual}
\end{figure}

\textbf{Metrics.}
Against the registered GT mesh, we report Chamfer distance, F-score, surface
IoU at $128^3$, and Sharp Normal Error (SNE) \citep{chen2025dora}, following
prior refinement and editing work
\citep{deng2024detailgen3d,li2025voxhammer,ye2025nano3d}.  On matched RGB views we report PSNR, SSIM\citep{wang2004ssim} and LPIPS \citep{zhang2018lpips}, as in 3D editing
benchmarks \citep{li2025voxhammer,ye2025nano3d}, plus DINOv2-large \citep{oquab2024dinov2} cosine
similarity (\emph{DINO}) and FID \citep{heusel2017fid}, following
\citet{xiang2024trellis} and \citet{li2025voxhammer}. We also use GPTEval3D
\citep{wu2024gpteval3d}: a vision-language judge (GPT-4o \citep{openai2024gpt4o} or GPT-5.6) compares
pairs of unlabeled assets on six criteria, and we report its 1--10 win share,
for which chance is $5.0$. Data splits, registration, and the metric, judging
and user-study protocols are in Appendix~\ref{app:eval}; the construction of
\bench{} is in Appendix~\ref{app:bigdetail}.

\begin{table}[t]
\centering
\begin{minipage}[t]{0.535\linewidth}
\centering
\caption{Human preference, TRELLIS.2 versus \mbox{\textsc{NOFACE3D}} on PAniC-3D
($16$ reviewers, $200$ assets). \emph{Neither} is a share of all responses;
the other rows are over judged ratings.}
\label{tab:user-study}
\small
\setlength{\tabcolsep}{2pt}
\begin{tabular*}{\linewidth}{@{\extracolsep{\fill}} l ccc}
\toprule
 & Joined & Genshin & VRoid \\
\midrule
Responses         & $1{,}188$ & $593$ & $595$ \\
Neither           & $12.0\%$ ($143$) & $6.4\%$ ($38$) & $17.6\%$ ($105$) \\
TRELLIS.2         & $12.2\%$ ($127$) & $9.0\%$ ($50$) & $15.7\%$ ($77$) \\
Tie               & $4.5\%$ ($47$) & $2.9\%$ ($16$) & $6.3\%$ ($31$) \\
\textsc{NOFACE3D} & $\mathbf{83.3\%}$ ($871$) & $\mathbf{88.1\%}$ ($489$)
                  & $\mathbf{78.0\%}$ ($382$) \\
\bottomrule
\end{tabular*}
\end{minipage}\hfill
\begin{minipage}[t]{0.445\linewidth}
\centering
\caption{GPTEval3D (GPT-5.6) on \mbox{\bench{}} ($N{=}339$), TRELLIS.2 versus
\mbox{\textsc{NOFACE3D}}. Head-to-head win share (1--10); each pair sums to $10$.}
\label{tab:gpteval-bigdetail}
\small
\setlength{\tabcolsep}{2pt}
\begin{tabular*}{\linewidth}{@{\extracolsep{\fill}} l cc}
\toprule
Criterion & TRELLIS.2 & \textsc{NOFACE3D} \\
\midrule
Fidelity              & $3.76$ & $\mathbf{6.24}$ \\
3D plausibility       & $4.60$ & $\mathbf{5.40}$ \\
Geo.--tex.\ alignment & $3.89$ & $\mathbf{6.11}$ \\
Texture detail        & $4.61$ & $\mathbf{5.39}$ \\
Geometry detail       & $4.21$ & $\mathbf{5.79}$ \\
Overall               & $4.26$ & $\mathbf{5.74}$ \\
\bottomrule
\end{tabular*}
\end{minipage}
\end{table}

\subsection{Main results}
\label{sec:baseline-results}

\textbf{Refinement and editing methods.}
\textsc{NOFACE3D} is best on nearly every metric on both datasets
(Table~\ref{tab:baselines}); TRELLIS.2 leads Chamfer on PAniC-3D by $0.001$.
DetailGen3D is best on SNE because it sharpens the whole surface, but this also
warps the shape (TexVerse Chamfer $35.4$ vs.\ $7.22$). Easy3E and Elevate3D fall
below TRELLIS.2 on most metrics.

Under GPT-5.6, \textsc{NOFACE3D} has the highest overall and geometry-detail
win shares among generated methods on both datasets (Table~\ref{tab:baselines}).
It even scores above GT on geometry detail ($8.07$ vs.\ $6.83$ on PAniC-3D and
$6.50$ vs.\ $6.33$ on TexVerse), since it produces denser geometric detail
consistent with the image. The gain is not detail density alone: DINO
similarity to GT also rises, from $0.792$ to $0.812$ on PAniC-3D and from
$0.728$ to $0.741$ on TexVerse.

In Figure~\ref{fig:qual}, \textsc{NOFACE3D} recovers the faces, hair and
accessories that TRELLIS.2 smears, and object detail such as the throne's side
panels, while leaving the rest of each asset unchanged. DetailGen3D warps the
throne and the telephone, Elevate3D blotches textures and breaks the telephone
cord, and Easy3E returns nearly blank character textures.

\begin{figure}[t]
\centering
\includegraphics[width=\linewidth]{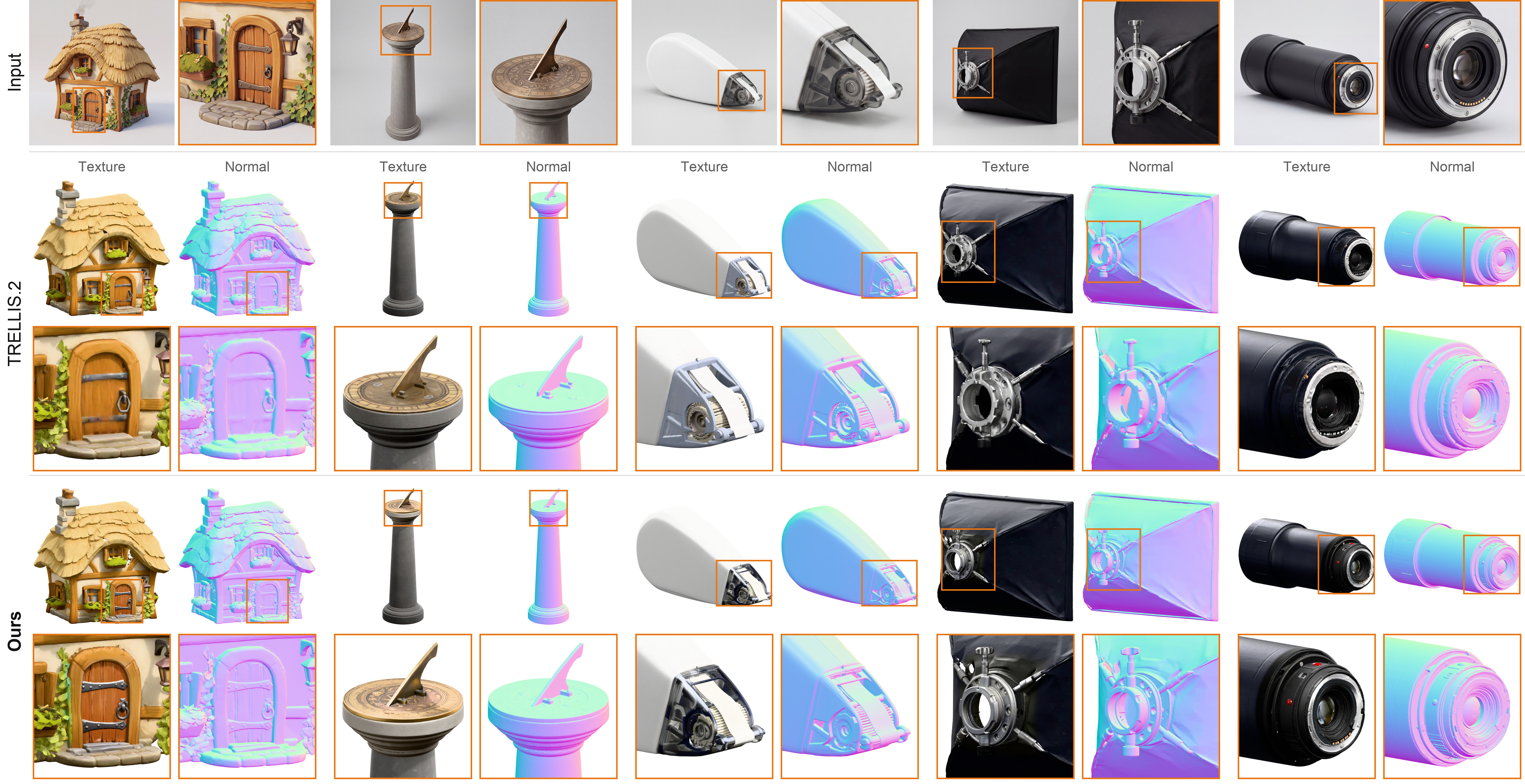}
\caption{\textbf{Qualitative results on \bench{}.} Top: input image with the
labeled region and its enlargement. Below: TRELLIS.2 and \textsc{NOFACE3D},
texture and normal map.}
\label{fig:bigdetail}
\end{figure}

\textbf{Human preference.}
\label{sec:user-study}
Sixteen reviewers compared TRELLIS.2 and \textsc{NOFACE3D} blind on $200$
PAniC-3D assets, giving $1{,}188$ responses on a $1$--$9$ scale. $83.3\%$ of
judged ratings prefer
\textsc{NOFACE3D} and $12.2\%$ TRELLIS.2, with a signed mean of $+1.83$ on
$[-4, +4]$ ($95\%$ CI $[+1.52, +2.11]$, reviewer-cluster bootstrap), on both
subsets (Table~\ref{tab:user-study}). $12.0\%$ of responses were
\emph{neither}, both generations too poor to judge; counting them,
$73.3\%$ of all responses prefer \textsc{NOFACE3D}.

\begin{figure}[t]
\centering
\includegraphics[width=0.6\linewidth]{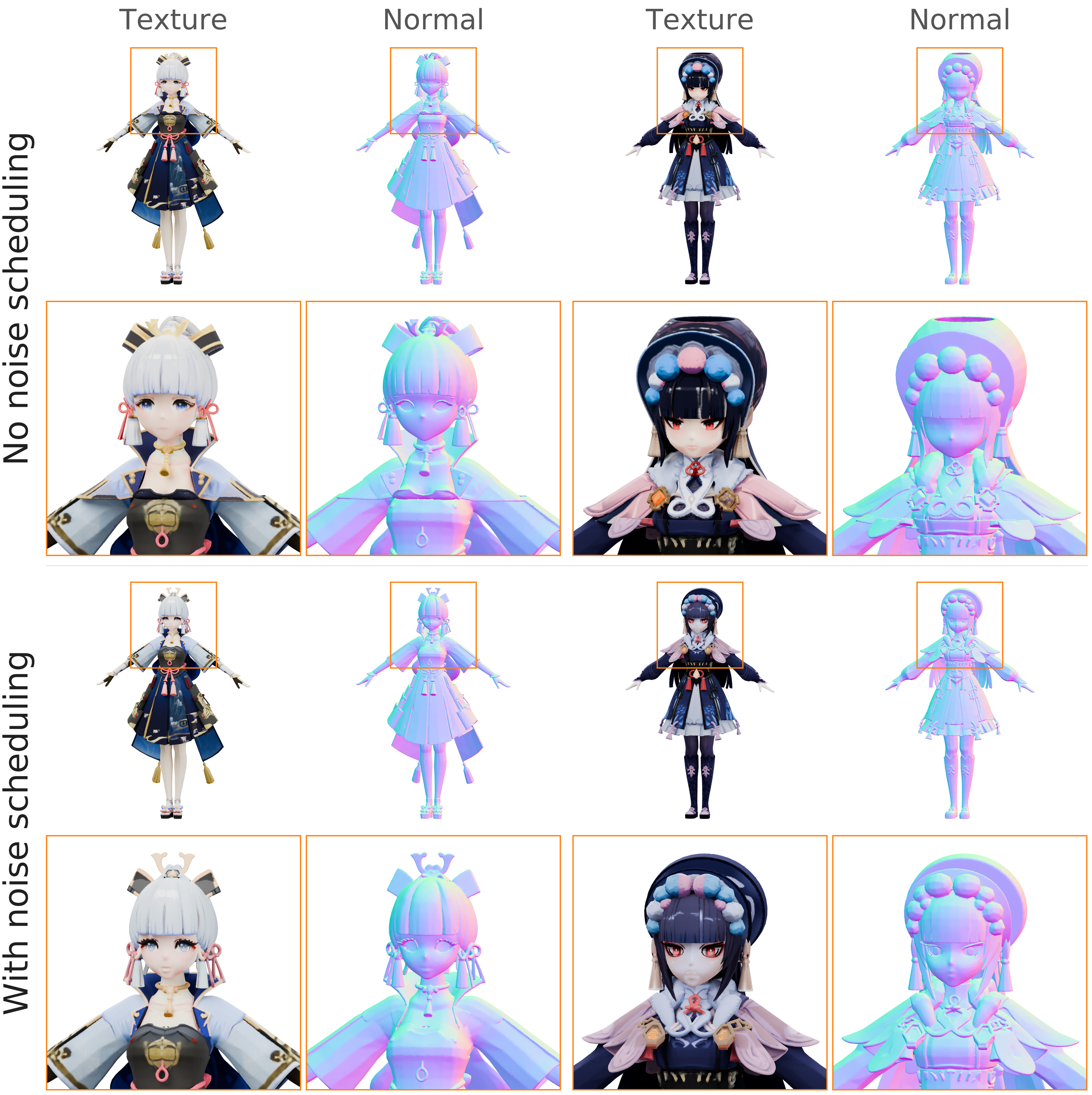}
\caption{\textbf{Noise scheduling ablation.} Top: regeneration without
anchoring or noise scheduling. Bottom: with noise scheduling. Texture and
normal views use matched cameras and detail crops. Orange boxes mark the
regions enlarged below each full-asset view.}
\label{fig:ablation-noise-scheduling}
\end{figure}

\textbf{Results on \bench{}.}
\label{sec:bigdetail-results}
On the $339$ refined \bench{} assets, GPT-5.6 prefers \textsc{NOFACE3D} to
TRELLIS.2 on all six criteria (Table~\ref{tab:gpteval-bigdetail}; overall
$5.74$ vs.\ $4.26$). Figure~\ref{fig:bigdetail} shows that \textsc{NOFACE3D} recovers details TRELLIS.2 blurs, such as the door straps, the tape-dispenser gear and the lens-mount contacts.

\begin{figure}[!htbp]
\centering
\includegraphics[width=0.5\linewidth]{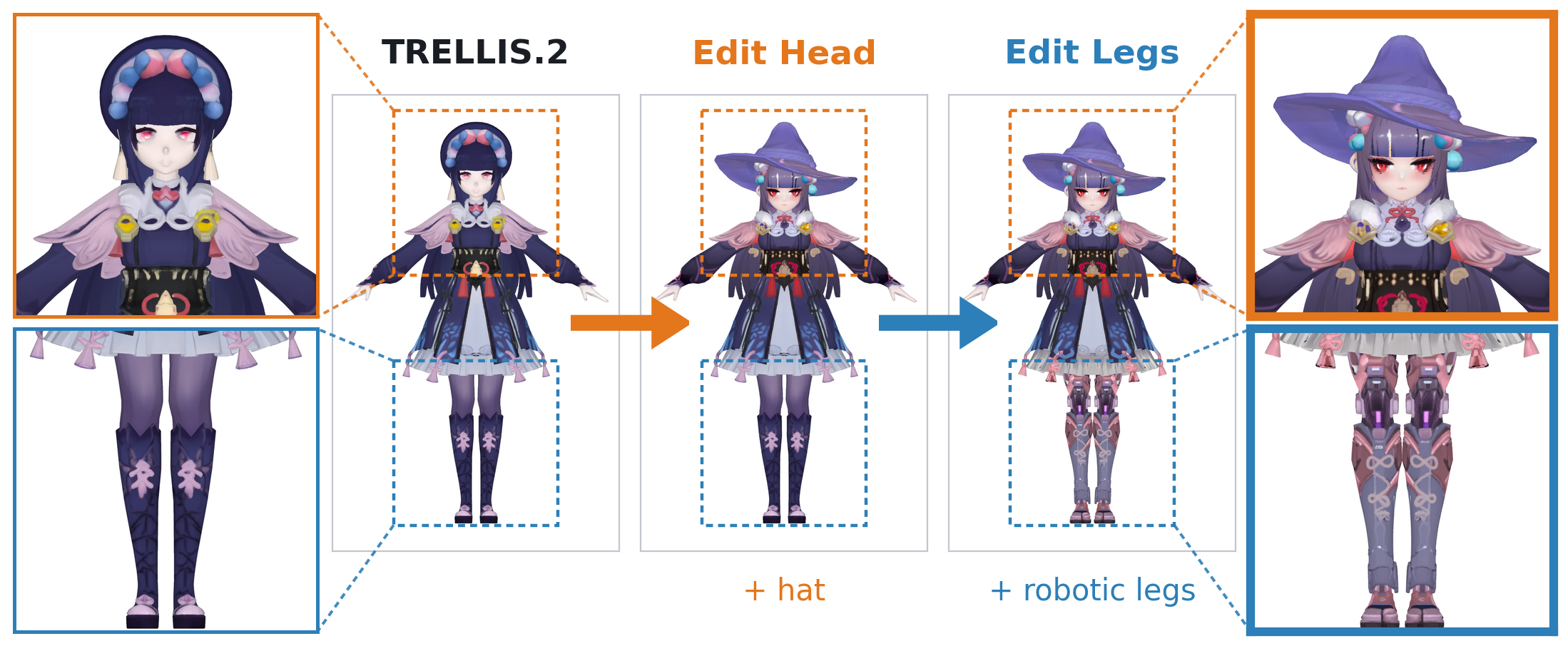}
\caption{\textbf{Localized 3D editing.} A Genshin character gains a hat (head
region), then robotic legs (leg region). Dashed boxes mark the regions.}
\label{fig:edit}
\end{figure}

\subsection{Noise scheduling ablation}
\label{sec:ablation}

Figure~\ref{fig:ablation-noise-scheduling} compares refinement with and
without noise scheduling. In the ablated variant, we regenerate the selected
region without boundary anchoring or the spatially varying noise schedule.
Regeneration alone does not reliably produce a smooth transition to the
retained asset, leaving visible discontinuities in geometry and appearance
at attachment boundaries. With noise scheduling, the regenerated region
connects more coherently to its surroundings.

\subsection{Other applications: localized 3D editing}
\label{sec:applications}

Conditioning on an edited crop turns \textsc{NOFACE3D} into a localized 3D
editor. We edit the input image with Gemini Flash Image and run the unchanged
refinement pipeline. Regeneration away from attachment boundaries allows
edits to change the region's shape, while the
anchored band and the mesh-space rejoin keep the rest of the asset close to the
original.
Edits compose: in Figure~\ref{fig:edit} a character gains a hat, then robotic
legs, and the second edit leaves the hat untouched. Please refer to our appendix for more examples.

\section{Conclusion}
\label{sec:conclusion}

We presented \textsc{NOFACE3D}, a training-free framework that refines the
detail-critical regions of an asset produced by a pretrained image-to-3D
generator. It reallocates the latent budget rather than enlarging it: the
detailed region's 3D content is rescaled to fill the generator's latent grid, re-encoded
there, and regenerated under a boundary-distance field that pins the
interface latents and restores them exactly, after which the refined region is
combined with the surrounding mesh. The method needs only the frozen generator,
the input image and a box around the target region, so it is not tied to a
category or body part, and conditioning the same process on an edited crop
turns it into a localized 3D editor.

\textbf{Limitations.}
Because \textsc{NOFACE3D} is training-free and runs on the frozen TRELLIS.2
generator, it inherits that generator's prior: re-encoding adds latent capacity
but does not change what the generator can produce, so text such as printed
labels and lettering, which TRELLIS.2 struggles to render legibly, often
remains distorted after refinement. Post-training the generator on text-rich
samples may alleviate this.

\bibliography{iclr2027_conference}
\bibliographystyle{iclr2027_conference}

\clearpage
\appendix
\section{Implementation details}
\label{app:impl}

\textbf{Generator.}
We use the released 4B TRELLIS.2 model in its default \texttt{1024\_cascade}
configuration. DINOv3 features are extracted at $512$ pixels for the first two
stages and at $1024$ pixels for the last two. The SS latent decodes to $64^3$
occupancy, max-pooled to the sparse $32^3$ support of the low-resolution shape
SLat; the shape decoder upsamples that SLat, and its predicted occupancy is
quantized to the sparse $64^3$ grid of the high-resolution shape SLat. The
texture SLat is sampled on the same $64^3$ coordinates with the shape SLat
concatenated to its input, and the texture decoder, guided by the shape
decoder's intermediate subdivisions, lifts it to six PBR channels (base color,
metallic, roughness, alpha). We use Euler sampling with time rescaling and
classifier-free guidance; the time ladders are specified below.

\textbf{Cost.}
With $\tau_{\max} = 1$, \textsc{NOFACE3D} samples each stage of the $1024^3$
cascade once per region with the released $12$-step ladder, so its shape stages
cost the same as in one TRELLIS.2 generation. \citet{xiang2025trellis2} report
about $17$\,s for a $1024^3$ asset on an NVIDIA H100, $10$\,s for shape and
$7$\,s for material. The only additional cost is voxelizing and encoding the crop seed, which uses the generator's own conversion routines and encoders. \textsc{NOFACE3D} loads the
same models as TRELLIS.2 and runs on the same hardware, which
\citet{xiang2025trellis2} specify as a GPU with at least $24$\,GB of memory.

\textbf{Crop box and condition.}
For PAniC-3D and TexVerse, we use square image bounding boxes and align the
base asset to the image by its frontal silhouette. Since a single image cannot
constrain depth, the crop box $B$ starts at the front-most surface under
$\Omega$ and its depth equals the image box's side length, retaining the
front-facing portion when the object extends farther back. This is an
experimental setting, not a requirement on the shape of the refinement box.
The condition crop passes through the
generator's standard preprocessing (background removal and recentering). In our
PAniC-3D and TexVerse experiments it is a crop of the input image and is not
upscaled, so no detail enters through an image model rather than through
\textsc{NOFACE3D}. The crop of a
user image can be low-resolution; it can then be upsampled by image
interpolation or by a vision-language image model such as Gemini Flash Image,
which touches only the 2D condition. \textsc{NOFACE3D} also runs on generated
inputs: the \bench{} images are generated with Gemini, and the editing crops of
Section~\ref{sec:applications} are Gemini edits.

\textbf{Coordinate transforms.}
Let $p$ denote parent-mesh coordinates and
$q=\phi_{\mathrm{o}}(p)$ coordinates in the box-aligned frame, where
$\phi_{\mathrm{o}}$ is the recorded alignment transform. For labeled oriented
boxes, it includes the box rotation and the inverse scale used to cache the
parent mesh. Let $b_{\min}$ and $b_{\max}$ be the selected submesh's tight
bounding-box corners in this frame. Its local center and uniform scale are
$c_\Omega = \tfrac12(b_{\min} + b_{\max})$ and
$s_\Omega = 0.99999/\lVert b_{\max} - b_{\min} \rVert_\infty$, giving local
coordinates $y=\phi_\Omega(q)=A\,s_\Omega(q-c_\Omega)$, where $A$ converts
Y-up coordinates to the generator's Z-up convention.
The complete mapping is $\phi_\Omega \circ \phi_{\mathrm{o}}$.
Local rescaling reduces the latent pitch in the box-aligned frame by a factor
of $s_\Omega$. Reintegration applies
$q=\phi_\Omega^{-1}(y)=A^{-1}y/s_\Omega+c_\Omega$, then
$p=\phi_{\mathrm{o}}^{-1}(q)$. Any initial parent-cache scaling is undone on
final export.

\textbf{Boundary-distance field.}
We write $\mathcal{G}$ for the nonempty set of attachment boundaries, identified once
per selection as
\begin{equation}
  \mathcal{G} = \big\{\, w : \text{a triangle of the base mesh straddles the
  plane of box face } w \text{ within its extent} \,\big\}.
  \label{eq:patches}
\end{equation}
This set can also be specified explicitly. A token at integer coordinate $n$ on a
stage's $R^3$ grid ($R=16$, $32$ or $64$) has normalized position
$u=(n+\tfrac12)/R\in[0,1]^3$, corresponding to the local cube $\mathcal{C}$
shifted to the unit cube. In this frame, each face $w$ of the crop box $B$ lies
in a plane $u_{a_w}=p_w$ perpendicular to axis $a_w$. Let $\ell_w$ denote the
box extent along that axis. The normalized inward distance is
\begin{equation}
  d(u) = \min_{w \in \mathcal{G}} d_w(u),
  \qquad
  d_w(u) = \pm\,\frac{u_{a_w} - p_w}{\ell_w},
  \label{eq:walldist}
\end{equation}
with $+$ for a lower face and $-$ for an upper face, so that $d_w$ is positive
inside $B$.
Because the smoothstep is monotone, applying Eq.~\ref{eq:field} to the smallest
normalized distance is equivalent to taking the minimum release level over
attachment boundaries. The band widths are
$(g, r) = (0.16, 0.04)$ on the SS latent, $(0.10, 0.06)$ on the shape SLats and
$(0.05, 0.05)$ on the texture SLat. They were chosen empirically and satisfy
Eq.~\ref{eq:nesting} with equality ($0.10+0.06=0.16$ and
$0.05+0.05=0.10$); both shape SLat stages use the same band. The release
maximum $\tau_{\max}$ is tunable independently at each stage. Lowering it on
SS retains more of the base structure at initialization, while the shape and
texture maxima separately control regeneration of geometry and appearance.

\textbf{Time ladders.}
Let $N$ be the configured step count and
$h_\alpha(t)=\alpha t/[1+(\alpha-1)t]$ the generator's time-rescaling function,
with rescale factor $\alpha$. For shape and texture SLats, the ladder starts
at $\tau_{\max}$ and keeps entries of
$\{h_\alpha(1-j/N)\}_{j=0}^{N}$ below $\tau_{\max}$, ending at zero.
The SS implementation instead uses $K=\max(1,\mathrm{round}(N\tau_{\max}))$
steps with $t_k=h_\alpha(\tau_{\max}(1-k/K))$. Thus its first model time need
not equal $\tau_{\max}$. All stages initialize tokens at their individual
$\tau_i$ and use Eq.~\ref{eq:anchor} thereafter: release levels are not an
upper bound on the noise encountered while anchored to the noised base latent.

\textbf{Stage outputs.}
A shape token whose coordinate has no counterpart in the base latent receives
a zero reference before standardization and $\tau_i=\tau_{\max}$.
After de-standardizing
$x(t_K)$ to $\hat{z}$, the low- and high-resolution shape SLat stages return
\begin{equation}
  z_i^{\mathrm{out}} = z_i + \lambda_i\, (\hat{z}_i - z_i),
  \qquad
  \lambda_i = \tau_i / \tau_{\max},
  \label{eq:blend}
\end{equation}
and on the final high-resolution shape SLat, tokens in the anchored band
are then overwritten with their base rows, coordinates
and features, so anchoring is enforced twice and independently, in the sampler
and in feature space. The SS latent is not blended; it is decoded to $64^3$
occupancy, every cell whose release level, upsampled from $16^3$ by nearest
neighbor, is at most $10^{-7}$ copies the base occupancy, and the result is
max-pooled to the $32^3$ support as in the base generator.

\textbf{Texture.}
The texture reference is aligned to the refined shape coordinates. Within the
\emph{shape} SLat's anchored band, it copies base texture values at matching
integer coordinates; elsewhere it is zero before standardization. This
provides base texture values throughout both the texture anchored band and its
transition band, since $g_{\mathrm{tex}}+r_{\mathrm{tex}} \le g_{\mathrm{shape}}$. The
standardized shape SLat is concatenated to the texture model's input as in the
base generator. The texture sampler uses the default settings of the released
TRELLIS.2 model \citep{xiang2025trellis2}: $12$ Euler steps with time-rescale
factor $3.0$, and guidance strength $1.0$ with guidance rescale $0$, which
disables classifier-free guidance.
The blend of Eq.~\ref{eq:blend} is omitted, so every released token keeps its sampled
value, and tokens in the anchored band are overwritten with their base rows as for the
shape SLat. The result decodes, guided by the refined shape decoder's
subdivisions, to the six PBR channels.

\textbf{Reintegration.}
Base vertices take PBR values queried from the base asset's PBR voxels, and
refined vertices take those of the refined decode. The full refined submesh
is concatenated with the retained parent faces; it is not clipped again to
the crop box.

\section{Evaluation details}
\label{app:eval}

\textbf{Data.}
We use the Genshin (PMX) and VRoid (VRM) subsets of PAniC-3D. On these
characters the head and upper body carry much of the perceived identity while
occupying little of the latent grid, which is why the marked region is a crop
of that part.
Of the $5{,}631$ PAniC-3D assets with completed original TRELLIS.2 results and
refinement ($132$ Genshin, $5{,}499$ VRoid), the $4{,}892$ that complete every
evaluator form the large-scale set, and the $329$ on which every baseline also
completed ($120$ Genshin, $209$ VRoid) form the comparison set. The original
TRELLIS.2 results are generated once and cached, so they differ from ours only
by the refinement. GT, original and refined meshes are registered and rendered
with a shared camera and normalization.

\textbf{Baselines.}
On PAniC-3D and TexVerse every method sees the same single image.
DetailGen3D outputs untextured meshes, so its appearance metrics are not
reported. Easy3E receives the image crop as its target view and the crop box as
its 3D mask.

\textbf{Metrics.}
Chamfer distance and F-score use $250{,}000$ points per mesh. SNE uses $22$
views on a Fibonacci sphere at $512^2$, and the appearance metrics use four
$1024^2$ views at azimuths $30^\circ$, $120^\circ$, $210^\circ$, $300^\circ$ and
pitch $30^\circ$. DINO and FID measure semantic and distributional similarity
rather than pixel-wise agreement. FID is computed over the set of rendered
views and has no per-asset deviation.

\textbf{Judged evaluation.}
Following GPTEval3D \citep{wu2024gpteval3d}, the judge is GPT-4o or GPT-5.6,
which we use as a scalable proxy for human preference; the user study of
Section~\ref{sec:user-study} agrees with it. The six criteria are fidelity and
completeness, 3D plausibility, geometry--texture alignment, texture detail,
geometry detail, and overall quality; the judge is not told which asset is
refined. The geometry--texture alignment criterion asks whether the geometric
detail is consistent with the image and texture, that is, how realistic the
geometry looks, not whether it matches GT. The judge sees four RGB views and
four normal maps of each asset at azimuths $0^\circ$, $90^\circ$, $180^\circ$,
$270^\circ$. In the \emph{pairwise} protocol it picks the better of two assets
or a tie; in the \emph{monadic} protocol it rates each asset from 1 to 10
alone, and pairwise outcomes are derived by comparing the independent ratings.
Both report the win share
$10\,(\text{wins} + 0.5\,\text{ties})/\text{games}$, for which chance is $5.0$.
Side-by-side judging is position-biased: in a pilot, GPT-4o chose the left
panel $68.5\%$ of the time. The comparison with baselines therefore counts both
orders of every pair, and the large-scale tournament ($1{,}000$ assets,
$8{,}000$ judge calls) keeps only decisions on which both orders agree. The
labeled trial
shows the artist mesh as a named reference and asks which of the two assets
matches it better; as in the tournament, only assets on which both orders agree
are counted.

\textbf{Human study.}
The study in Section~\ref{sec:user-study} pairs TRELLIS.2 and
\textsc{NOFACE3D} on $200$ PAniC-3D assets from the Genshin and VRoid subsets.
Each trial shows rotating textured and normal-map views of both assets and the
conditioning image of the full object and of the marked region; the options
are unlabeled, with left/right assignment and orientation randomized. Reviewers
rate $1$--$9$ (left to right preference) or answer \emph{neither} when the base
generation is too poor to judge. Ratings are mapped so that $9$ favours
\textsc{NOFACE3D}, and the signed score is the mapped rating minus $5$.
Sixteen reviewers gave $1{,}188$ responses ($7$--$181$ judged ratings each,
median $35$); no demographic information was collected.

\textbf{Human study statistics.}
Signed mean $+1.83$ (s.d.\ $1.76$) over judged ratings, with $95\%$ CIs
$[+1.52, +2.11]$ (reviewer-cluster bootstrap), $[+1.64, +2.00]$ (asset-cluster
bootstrap) and $[+1.32, +1.87]$ around $+1.59$ (crossed reviewer--asset
random-effects model; variance $48\%$ asset, $4\%$ reviewer, $48\%$ residual).
Krippendorff's $\alpha$ \citep{krippendorff2018content} is $0.44$ on signed scores and $0.42$ on the four
labels. \emph{Neither} is $12.0\%$ of responses ($6.4\%$ Genshin, $17.6\%$
VRoid); counting it, shares are TRELLIS.2 $10.7\%$, tie $4.0\%$,
\textsc{NOFACE3D} $73.3\%$, and coding it as a tie gives a mean of $+1.61$.

\section{\bench{} construction}
\label{app:bigdetail}

TRELLIS.2 already reconstructs Objaverse-style assets well, such assets may
overlap its training data, and few of them concentrate their detail in a small
volume that a uniform latent grid under-resolves. Every \bench{} subject is
designed around that case. The images are generated, so no artist mesh exists
and none of the reference-based metrics can be computed; \bench{} is evaluated
with pairwise GPTEval3D on the whole asset, the original TRELLIS.2 result
against the refined asset for the same input.

\textbf{Subjects and images.}
Each subject is specified by an object, its \emph{smooth mass} (e.g.\ ``red
cylindrical steel tank''), and a compact \emph{detail cluster} with named parts
(e.g.\ ``valve and trigger assembly: squeeze lever, carry handle, pressure
gauge, safety pin, tamper seal''). Subjects span everyday and industrial
objects: stationery and office equipment, hand and power tools, kitchen and
household appliances, furniture and lighting, musical instruments, sports gear,
cameras and laboratory instruments, vehicles, boats and aircraft, railway,
firefighting, utility and space hardware, and buildings. For each subject we
generate one $4096{\times}4096$ image with Nano Banana 2
(\texttt{gemini-3.1-flash-image}) and reconstruct it with the original
TRELLIS.2 model.

\textbf{Region labels.}
Annotators mark each detail-critical region as a pair of boxes, a rectangle on
the image and an oriented 3D box on the mesh, with no assumed camera; an asset
may carry several regions, or be marked as needing no refinement and excluded.
The evaluation set is the $339$ refined assets.

\textbf{Refinement.}
We use the annotated bounding boxes directly:
the 3D box specifies the selection and its orientation, whole triangles whose centroids
fall inside it are kept, and the attachment boundaries $\mathcal{G}$ are the
box faces crossed by the base mesh, as in Eq.~\ref{eq:patches}. The per-stage
band widths are those of Appendix~\ref{app:impl}, with independently
configurable release maxima. The condition is the exact image crop, resized
with Lanczos interpolation and padded to a white $1024^2$ square before the
generator's background removal, without generative image sharpening. Each labeled region is refined
in its own run.

\section{Large-scale comparison with TRELLIS.2}
\label{app:scale}

Table~\ref{tab:app-scale} compares \textsc{NOFACE3D} with the original TRELLIS.2
results on all $4{,}892$ PAniC-3D assets. Every appearance metric improves,
most clearly FID ($13.97 \to 7.43$) and DINO ($0.812 \to 0.830$), the fine
F-score and surface IoU rise (F1 at $0.25\%$ from $0.141$ to $0.144$, IoU from
$0.373$ to $0.381$), and pointwise geometry otherwise changes little: Chamfer
differs by $0.33\%$, F1 at $1\%$ by less than $0.1\%$ and SNE by $0.2\%$. This
is consistent with the rest of the asset being preserved.
The refinement shows in appearance against
GT and in judged and human preference.

Under the monadic protocol, both judges prefer \textsc{NOFACE3D} to TRELLIS.2 on
all six criteria, with overall quality $4.81$ vs.\ $4.22$ on GPT-4o and $5.02$
vs.\ $3.53$ on GPT-5.6 (Table~\ref{tab:app-judge-monadic}); the margins are
moderate because most of each asset is the preserved remainder. A pairwise
tournament on a $1{,}000$-asset subset, keeping only decisions on which both
orders agree, gives the same ordering. Finally, a labeled trial shows the
artist mesh as a named reference and asks which of the two matches it better:
of the $871$ assets on which both orders agree, the judge chose \textsc{NOFACE3D}
on $681$ and TRELLIS.2 on $188$, $78.4\%$ of the $869$ decided assets, and
judged the remaining $2$ a tie.

\begin{table}[ht]
\centering
\caption{TRELLIS.2 versus \textsc{NOFACE3D} on $4{,}892$ PAniC-3D assets against
artist GT. Mean $\pm$ population s.d.; Chamfer and SNE ${\times}100$; FID is
set-level.}
\label{tab:app-scale}
\small
\begin{tabular*}{\textwidth}{@{\extracolsep{\fill}} l c c}
\toprule
Metric & TRELLIS.2 & \textsc{NOFACE3D} \\
\midrule
Chamfer $\downarrow$     & $\mathbf{3.359 \pm 2.764}$   & $3.370 \pm 2.773$ \\
F1 @ $0.25\%$ $\uparrow$ & $0.141 \pm 0.070$            & $\mathbf{0.144 \pm 0.070}$ \\
F1 @ $1\%$ $\uparrow$    & $0.37219 \pm 0.15536$        & $0.37222 \pm 0.15548$ \\
Surface IoU $128^3$ $\uparrow$
                         & $0.373 \pm 0.140$            & $\mathbf{0.381 \pm 0.142}$ \\
SNE $\downarrow$         & $\mathbf{1.312 \pm 0.610}$   & $1.314 \pm 0.620$ \\
PSNR $\uparrow$          & $15.70 \pm 2.17$             & $\mathbf{16.53 \pm 2.76}$ \\
SSIM $\uparrow$          & $0.8884 \pm 0.0290$          & $\mathbf{0.8906 \pm 0.0299}$ \\
LPIPS $\downarrow$       & $0.1163 \pm 0.0265$          & $\mathbf{0.1123 \pm 0.0268}$ \\
DINO $\uparrow$          & $0.812 \pm 0.114$            & $\mathbf{0.830 \pm 0.119}$ \\
FID $\downarrow$         & $13.97$                      & $\mathbf{7.43}$ \\
\bottomrule
\end{tabular*}
\end{table}

\begin{table}[ht]
\centering
\caption{Monadic judged comparison on $4{,}892$ PAniC-3D assets: each mesh is
rated alone, and win shares (1--10) are derived from the ratings.}
\label{tab:app-judge-monadic}
\small
\setlength{\tabcolsep}{3pt}
\begin{tabular*}{\textwidth}{@{\extracolsep{\fill}} l ccc ccc}
\toprule
& \multicolumn{3}{c}{GPT-4o} & \multicolumn{3}{c}{GPT-5.6} \\
\cmidrule(lr){2-4} \cmidrule(lr){5-7}
Criterion & GT & TRELLIS.2 & \textsc{NOFACE3D} & GT & TRELLIS.2 & \textsc{NOFACE3D} \\
\midrule
Fidelity / completeness & $5.68$ & $4.41$ & $\mathbf{4.91}$ & $6.38$ & $3.61$ & $\mathbf{5.01}$ \\
3D plausibility         & $5.63$ & $4.64$ & $\mathbf{4.73}$ & $5.87$ & $4.33$ & $\mathbf{4.80}$ \\
Geo.--tex.\ alignment   & $5.74$ & $4.44$ & $\mathbf{4.82}$ & $4.94$ & $4.67$ & $\mathbf{5.40}$ \\
Texture details         & $6.05$ & $4.08$ & $\mathbf{4.86}$ & $7.46$ & $2.42$ & $\mathbf{5.12}$ \\
Geometry details        & $5.64$ & $4.63$ & $\mathbf{4.73}$ & $5.29$ & $4.75$ & $\mathbf{4.96}$ \\
Overall quality         & $5.97$ & $4.22$ & $\mathbf{4.81}$ & $6.45$ & $3.53$ & $\mathbf{5.02}$ \\
\bottomrule
\end{tabular*}
\end{table}

\section{Additional results}
\label{app:results}

Tables~\ref{tab:app-panic} and~\ref{tab:app-texverse} repeat the comparison
with refinement and editing methods with standard deviations across assets,
and Tables~\ref{tab:app-judge-panic} and~\ref{tab:app-judge-texverse} give its
judged results on all six criteria. Table~\ref{tab:app-judge-bigdetail} gives
both judges' scores on \bench{}. Figure~\ref{fig:app-edit} shows two more editing
sequences.

\medskip
\noindent\begin{minipage}{\linewidth}
\centering
\begin{minipage}[c]{0.49\linewidth}
  \includegraphics[width=\linewidth]{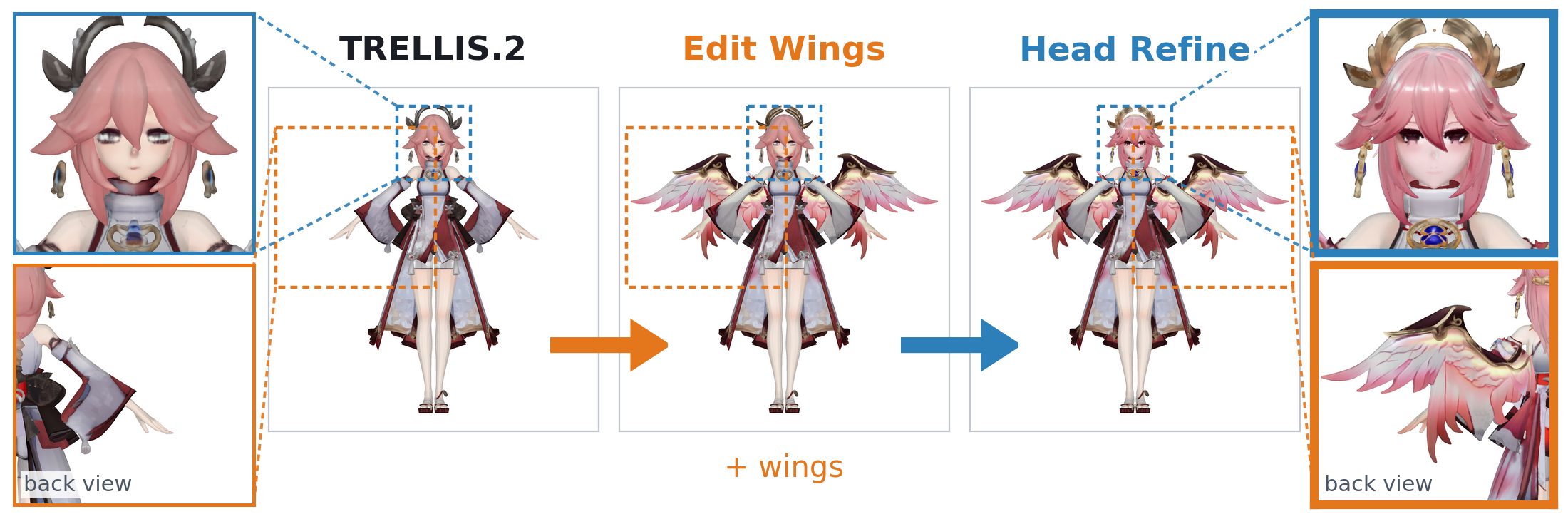}
\end{minipage}\hfill
\textcolor{black!25}{\rule[-0.08\linewidth]{0.4pt}{0.16\linewidth}}\hfill
\begin{minipage}[c]{0.49\linewidth}
  \includegraphics[width=\linewidth]{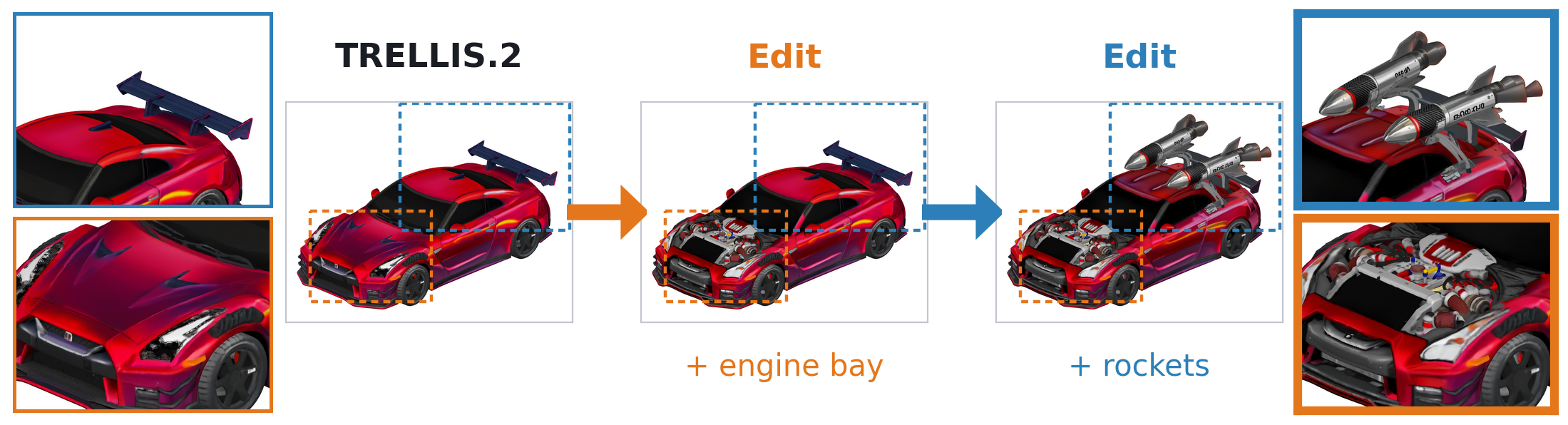}
\end{minipage}
\captionof{figure}{\textbf{More editing sequences.} Left: a Genshin character
gains wings, then its head is refined. Right: a car gains an engine bay (front),
then rockets (rear).}
\label{fig:app-edit}
\end{minipage}
\medskip

\begin{table}[ht]
\centering
\caption{Objective comparison on PAniC-3D ($N{=}329$). Mean $\pm$ s.d.; Chamfer
and SNE ${\times}100$; FID is set-level.}
\label{tab:app-panic}
\footnotesize
\setlength{\tabcolsep}{2.5pt}
\begin{tabular*}{\textwidth}{@{\extracolsep{\fill}} l ccccc}
\toprule
Metric & TRELLIS.2 & Easy3E & Elevate3D & DetailGen3D & \textsc{NOFACE3D} \\
\midrule
Chamfer $\downarrow$
  & $\mathbf{2.938 \pm 1.884}$ & $3.955 \pm 2.071$ & $3.059 \pm 1.896$
  & $2.939 \pm 1.900$ & $2.939 \pm 1.878$ \\
F1 @ $0.25\%$ $\uparrow$
  & $0.138 \pm 0.073$ & $0.074 \pm 0.034$ & $0.115 \pm 0.051$
  & $0.127 \pm 0.056$ & $\mathbf{0.142 \pm 0.074}$ \\
F1 @ $1\%$ $\uparrow$
  & $0.389 \pm 0.166$ & $0.265 \pm 0.111$ & $0.372 \pm 0.146$
  & $\mathbf{0.393 \pm 0.153}$ & $0.389 \pm 0.165$ \\
Surface IoU $128^3$ $\uparrow$
  & $0.383 \pm 0.144$ & $0.294 \pm 0.113$ & $0.377 \pm 0.135$
  & $0.386 \pm 0.141$ & $\mathbf{0.391 \pm 0.147}$ \\
SNE $\downarrow$
  & $1.12 \pm 0.40$ & $1.24 \pm 0.37$ & $1.26 \pm 0.38$
  & $\mathbf{1.04 \pm 0.40}$ & $1.12 \pm 0.40$ \\
PSNR $\uparrow$
  & $15.80 \pm 1.90$ & $13.17 \pm 1.08$ & $16.59 \pm 1.76$ & -- & $\mathbf{16.78 \pm 2.44}$ \\
SSIM $\uparrow$
  & $0.882 \pm 0.023$ & $0.878 \pm 0.023$ & $0.876 \pm 0.021$ & -- & $\mathbf{0.883 \pm 0.025}$ \\
DINO $\uparrow$
  & $0.792 \pm 0.097$ & $0.588 \pm 0.094$ & $0.671 \pm 0.083$ & -- & $\mathbf{0.812 \pm 0.103}$ \\
LPIPS $\downarrow$
  & $0.118 \pm 0.021$ & $0.128 \pm 0.019$ & $0.122 \pm 0.018$ & -- & $\mathbf{0.115 \pm 0.021}$ \\
FID $\downarrow$
  & $43.17$ & $98.74$ & $68.25$ & -- & $\mathbf{33.02}$ \\
\bottomrule
\end{tabular*}
\end{table}

\begin{table}[ht]
\centering
\caption{Objective comparison on TexVerse ($N{=}151$). Mean $\pm$ s.d.; Chamfer
and SNE ${\times}100$.}
\label{tab:app-texverse}
\footnotesize
\setlength{\tabcolsep}{2.5pt}
\begin{tabular*}{\textwidth}{@{\extracolsep{\fill}} l ccccc}
\toprule
Metric & TRELLIS.2 & Easy3E & Elevate3D & DetailGen3D & \textsc{NOFACE3D} \\
\midrule
Chamfer $\downarrow$
  & $7.27 \pm 14.9$ & $11.7 \pm 19.3$ & $13.7 \pm 18.7$
  & $35.4 \pm 31.0$ & $\mathbf{7.22 \pm 14.8}$ \\
F1 @ $0.25\%$ $\uparrow$
  & $0.123 \pm 0.147$ & $0.060 \pm 0.059$ & $0.058 \pm 0.064$
  & $0.009 \pm 0.010$ & $\mathbf{0.140 \pm 0.150}$ \\
F1 @ $1\%$ $\uparrow$
  & $0.306 \pm 0.230$ & $0.219 \pm 0.188$ & $0.194 \pm 0.178$
  & $0.032 \pm 0.031$ & $\mathbf{0.320 \pm 0.234}$ \\
Surface IoU $128^3$ $\uparrow$
  & $0.334 \pm 0.247$ & $0.250 \pm 0.193$ & $0.198 \pm 0.169$
  & $0.055 \pm 0.044$ & $\mathbf{0.337 \pm 0.244}$ \\
SNE $\downarrow$
  & $1.97 \pm 1.33$ & $1.81 \pm 1.37$ & $2.16 \pm 1.40$
  & $\mathbf{1.23 \pm 0.84}$ & $1.61 \pm 1.18$ \\
PSNR $\uparrow$
  & $18.02 \pm 9.24$ & $17.25 \pm 9.10$ & $16.97 \pm 9.15$ & -- & $\mathbf{19.57 \pm 9.29}$ \\
SSIM $\uparrow$
  & $0.860 \pm 0.066$ & $0.851 \pm 0.068$ & $0.836 \pm 0.071$ & -- & $\mathbf{0.863 \pm 0.065}$ \\
DINO $\uparrow$
  & $0.728 \pm 0.134$ & $0.663 \pm 0.139$ & $0.620 \pm 0.133$ & -- & $\mathbf{0.741 \pm 0.130}$ \\
LPIPS $\downarrow$
  & $0.135 \pm 0.061$ & $0.145 \pm 0.063$ & $0.158 \pm 0.066$ & -- & $\mathbf{0.134 \pm 0.061}$ \\
  FID $\downarrow$
  & $66.56$ & $74.87$ & $73.94$ & -- & $\mathbf{64.92}$ \\
\bottomrule
\end{tabular*}
\end{table}

\begin{table}[ht]
\centering
\caption{Pairwise judged comparison on PAniC-3D ($N{=}329$). Win share (1--10,
chance $5.0$).}
\label{tab:app-judge-panic}
\small
\begin{tabular*}{\textwidth}{@{\extracolsep{\fill}} l cccccc}
\toprule
Criterion & GT & TRELLIS.2 & Easy3E & Elevate3D & DetailGen3D & \textsc{NOFACE3D} \\
\midrule
\multicolumn{7}{l}{\emph{GPT-4o}} \\
Fidelity / completeness & $7.84$ & $6.41$ & $2.17$ & $4.73$ & -- & $\mathbf{7.57}$ \\
3D plausibility         & $7.17$ & $5.77$ & $2.76$ & $2.96$ & $4.54$ & $\mathbf{6.81}$ \\
Geo.--tex.\ alignment   & $7.90$ & $6.45$ & $2.41$ & $4.39$ & -- & $\mathbf{7.60}$ \\
Texture details         & $7.87$ & $6.46$ & $1.95$ & $4.93$ & -- & $\mathbf{7.61}$ \\
Geometry details        & $7.36$ & $5.88$ & $2.23$ & $3.04$ & $4.50$ & $\mathbf{6.98}$ \\
Overall quality         & $7.87$ & $6.43$ & $2.32$ & $4.49$ & -- & $\mathbf{7.59}$ \\
\midrule
\multicolumn{7}{l}{\emph{GPT-5.6}} \\
Fidelity / completeness & $8.52$ & $5.97$ & $1.80$ & $5.14$ & -- & $\mathbf{8.20}$ \\
3D plausibility         & $7.74$ & $6.56$ & $3.38$ & $0.69$ & $4.27$ & $\mathbf{7.36}$ \\
Geo.--tex.\ alignment   & $7.49$ & $7.00$ & $3.37$ & $3.33$ & -- & $\mathbf{8.53}$ \\
Texture details         & $9.05$ & $5.91$ & $2.03$ & $4.83$ & -- & $\mathbf{8.11}$ \\
Geometry details        & $6.83$ & $6.82$ & $3.02$ & $0.71$ & $4.55$ & $\mathbf{8.07}$ \\
Overall quality         & $8.52$ & $6.48$ & $2.44$ & $3.81$ & -- & $\mathbf{8.36}$ \\
\bottomrule
\end{tabular*}
\end{table}

\begin{table}[ht]
\centering
\caption{Pairwise judged comparison on TexVerse ($N{=}151$). Win share (1--10,
chance $5.0$).}
\label{tab:app-judge-texverse}
\small
\begin{tabular*}{\textwidth}{@{\extracolsep{\fill}} l cccccc}
\toprule
Criterion & GT & TRELLIS.2 & Easy3E & Elevate3D & DetailGen3D & \textsc{NOFACE3D} \\
\midrule
\multicolumn{7}{l}{\emph{GPT-4o}} \\
Fidelity / completeness & $5.58$ & $5.24$ & $5.34$ & $5.07$ & -- & $\mathbf{5.72}$ \\
3D plausibility         & $5.61$ & $5.21$ & $5.41$ & $5.01$ & $3.12$ & $\mathbf{5.64}$ \\
Geo.--tex.\ alignment   & $5.59$ & $5.23$ & $5.38$ & $5.05$ & -- & $\mathbf{5.66}$ \\
Texture details         & $5.62$ & $5.25$ & $5.34$ & $5.03$ & -- & $\mathbf{5.71}$ \\
Geometry details        & $5.58$ & $5.22$ & $5.39$ & $5.04$ & $3.12$ & $\mathbf{5.65}$ \\
Overall quality         & $5.59$ & $5.22$ & $5.37$ & $5.04$ & -- & $\mathbf{5.66}$ \\
\midrule
\multicolumn{7}{l}{\emph{GPT-5.6}} \\
Fidelity / completeness & $7.28$ & $4.05$ & $5.75$ & $3.82$ & -- & $\mathbf{7.08}$ \\
3D plausibility         & $7.01$ & $3.10$ & $\mathbf{6.26}$ & $2.89$ & $4.53$ & $6.22$ \\
Geo.--tex.\ alignment   & $7.05$ & $4.14$ & $6.25$ & $4.45$ & -- & $\mathbf{7.31}$ \\
Texture details         & $7.92$ & $4.37$ & $5.67$ & $4.73$ & -- & $\mathbf{7.14}$ \\
Geometry details        & $6.33$ & $3.09$ & $5.61$ & $3.14$ & $5.32$ & $\mathbf{6.50}$ \\
Overall quality         & $7.30$ & $3.84$ & $6.13$ & $3.89$ & -- & $\mathbf{7.22}$ \\
\bottomrule
\end{tabular*}
\end{table}

\begin{table}[ht]
\centering
\caption{Pairwise judged comparison on \bench{} ($N{=}339$), TRELLIS.2 versus
\textsc{NOFACE3D}. Head-to-head win share (1--10); each pair sums to $10$.}
\label{tab:app-judge-bigdetail}
\small
\begin{tabular*}{\textwidth}{@{\extracolsep{\fill}} l cc cc}
\toprule
& \multicolumn{2}{c}{GPT-4o} & \multicolumn{2}{c}{GPT-5.6} \\
\cmidrule(lr){2-3} \cmidrule(lr){4-5}
Criterion & TRELLIS.2 & \textsc{NOFACE3D} & TRELLIS.2 & \textsc{NOFACE3D} \\
\midrule
Fidelity / completeness & $4.40$ & $\mathbf{5.60}$ & $3.76$ & $\mathbf{6.24}$ \\
3D plausibility         & $4.42$ & $\mathbf{5.58}$ & $4.60$ & $\mathbf{5.40}$ \\
Geo.--tex.\ alignment   & $4.40$ & $\mathbf{5.60}$ & $3.89$ & $\mathbf{6.11}$ \\
Texture details         & $4.57$ & $\mathbf{5.43}$ & $4.61$ & $\mathbf{5.39}$ \\
Geometry details        & $4.60$ & $\mathbf{5.40}$ & $4.21$ & $\mathbf{5.79}$ \\
Overall quality         & $4.58$ & $\mathbf{5.42}$ & $4.26$ & $\mathbf{5.74}$ \\
\bottomrule
\end{tabular*}
\end{table}

\end{document}